\documentclass{article}

     \usepackage[final]{neurips_2019}

\usepackage[utf8]{inputenc} 
\usepackage[T1]{fontenc}    
\usepackage{hyperref}       
\usepackage{url}            
\usepackage{booktabs}       
\usepackage{amsfonts}       
\usepackage{nicefrac}       
\usepackage{microtype}      
\usepackage{graphicx}
\usepackage{titlesec}
\usepackage{makecell}

\title{Learning video embedding space with Natural Language Supervision}

\begin{document}

%

  
  

  

\noindent\rule{\textwidth}{4pt}

   \begin{center}
      \Large\textbf{Learning video embedding space with Natural Language Supervision}\\
      \noindent\rule{\textwidth}{1pt}

      \vspace{0.5cm}
      \normalsize\textbf{Phani Krishna Uppala\space\space\space\space\space\space\space\space\space\space\space\space\space\space 
      Abhishek Bamotra \space\space\space\space\space\space\space
      Shriti Priya\space\space\space\space\space\space\space\space\space\space\space\space\space\space Vaidehi Joshi}
   \end{center}

\begin{abstract}

The recent success of the CLIP model has shown its potential to be applied to a wide range of vision and language tasks. However this only establishes embedding space relationship of language to images, not to the video domain.  In this paper, we propose a novel approach to map video embedding space to natural langugage. We propose a two-stage approach that first extracts visual features from each frame of a video using a pre-trained CNN, and then uses the CLIP model to encode the visual features for the video domain, along with the corresponding text descriptions. We evaluate our method on two benchmark datasets, UCF101 and HMDB51, and achieve state-of-the-art performance on both tasks.

\end{abstract}

\section{Introduction}

Text query based image retrieval systems have recently shown drastic improvement \cite{radford2021learning, lei2021more, Uppala-2021-129108}. These approaches compute textual embedding for the given sentence using a language models like GPT-3 \cite{brown2020language} and compute visual embedding using a pre-trained Imagenet \cite{imagenet_cvpr09}. It is then trained for a matching function, which can be thought of as a learned similarity (ie. weighted - dot product). Since we are using dot product to find the maximum match, for any given text embedding there will exist a sub-space of visual embedding which maximizes the dot product. Do all samples in this visual embedding sub-space correspond to the coherent images? Or do some samples in this sub-space form an adversarial image? We will explore this direction with a goal of decoding the visual embedding space of a given text query embedding. Further, consecutive images in a video have both similar semantic and pixel-value distribution. Does this translate to videos being represented by continuous path walks in the visual embedding sub-space?

In order to answer the above questions, we want to use multiple tools including visualizations. Visualizations reveal what the network is looking at. For a given sentence embedding, we first find sub-space of visual embedding that match with the sentence embedding. Using these visual embedding and visualization techniques, we find the images that will produce a visual embedding corresponding to the textual embedding within the sub-space of interest.

Content Based Video Retrieval [egs: YouTube] is desirable because searches that rely purely on metadata are dependent on annotation quality and completeness. The presented architecture provides an efficient way of Text based Video Retrieval.
Also, our work is an extension of CLIP Model Action Recognition [RN50x16], where they have used Mid-frame level approach to get an accuracy of 53.4\%, while our method gives an improvement of 83.9\% [LSTM] and 85.5\% [Transformer] comparing to the SOTA on Kinetics 400 dataset 84.8\%.

State of the art computer vision methods \cite{8543868, kundu2020unsupervised} are usually structured for a fixed set of output categories or a fixed training data distribution on which the model is trained on.
This imposes a restricted form of supervision on learning. This restriction indicates that the model is only going to recognize and classify into one of the output classes on which it has learned on or on which it has been trained on. This form of restricted supervision does not prove to be helpful in a real world scenario wherein there could be visual concepts from all sorts of classes on which the model has not been trained on previously.
Thus, the restricted form of supervision can often limit the generality and usability of the SOTA methods for performing well on other visual concepts. In such a scenario, in order to learn new visual concepts, some form of context can help the model to perform a valid inference at test time and thus learn new visual concepts simultaneously with some help from the context information. With respect to context, zero-shot transfer, natural language supervision and multimodal learning are capable of providing this extra information or context required for the model to learn from and to accurately infer from.
Open set recognition also proves to be helpful and has been proven to work well with real life examples or scenarios. In open set recognition, incomplete knowledge of the world is present at training time and unknown classes can be submitted to an algorithm during testing. Open set recognition handles these unknown and unseen classes of images efficiently at test time based on certain statistical modelling. We aim to explore more into the Open set recognition approach through our project.
\section{Related Works}
\subsection{Open Set}
Deep neural networks have made breakthroughs in a wide range of visual understanding tasks. A typical challenge that hinders their real-world applications is that unknown samples may be fed into the system during the inference phase, but traditional deep neural networks or SOTA computer vision methods will wrongly recognize these unknown samples as one of the known classes. Open set recognition (OSR) is a potential solution to overcome this problem, where the open set classifier should have the flexibility to handle unknown samples and meanwhile maintain high classification accuracy in known classes.
Consequently, there has been a lot of relevant prior work in this area. Open Set Recognition with Conditional Probabilistic Generative Models \cite{imagenet_cvpr09, Mopuri_2018_ECCV} proposes an open set classifier which has the flexibility to reject unknown samples posed to the model at test time. In this paper, a novel framework, called Conditional Probabilistic Generative Models (CPGM), for open set recognition is proposed. The core insight of this work is to add discriminative information into the probabilistic generative models, such that the proposed models can not only detect unknown samples but also classify known classes by forcing different latent features to approximate conditional Gaussian distributions.
Unified Probabilistic Deep Continual Learning through Generative Replay and Open Set Recognition \cite{feichtenhofer2019slowfast} introduces a probabilistic approach to unify open set recognition with the prevention of catastrophic forgetting in deep continual learning, based on variational Bayesian inference\cite{Kundu_2018_CVPR}. In order to successfully distinguish unseen unknown data from trained known tasks, the paper proposes to bound the class specific approximate posterior by fitting regions of high density on the basis of correctly classified data points.
These bounds are further used to significantly alleviate catastrophic forgetting by avoiding samples from low density areas in generative replay. There is much more prior work which involves rejecting or avoiding the unseen or unknown input samples (that are out of the training data distribution or are from a different output class category) at inference time.

We aim to propose an approach where we perform an Open Set Recognition (OSR) on the unknown or unseen samples but instead of avoiding or rejecting them, we find a way to automatically handle these samples at the inference time. In order to explore this approach more in the open set conditions, we also aim to perform experiments for Video Retrieval from textual input using an image based CLIP model, but on a Video input dataset.

\subsection{Image Retrieval}
The neural network will be trained on text and visual embedding pairs to make it learn a mapping between text and corresponding images\cite{8658966}. Based on a supervised learning approach, labels, Image and text embedding will be used to match the unseen text data examples to a corresponding image. Likewise, we will achieve Image retrieval.

\subsection{Neuron visualizations}
As mentioned in the Introduction, we aim to use visualizations as a tool to understand what exactly a network is looking at. For this purpose, we need to find a match or correlation between the sentence embedding and a certain subspace of visual embedding\cite{swain2016study, uppala2023dynamic}. We propose to do so by using different neuron visualization techniques. Some of them include visualizing receptive fields or by visualizing activation through backpropagation. Both of these can lead us to finding the subspace of visual embedding that we are looking for. 
Visual embedding subspace or visual explanations of what a network sees at every stage of the learning process can also be achieved using Gradient-Weighted Class Activation Mapping \cite{Selvaraju_2019} which can prove in extrapolating semantic sense from the network's learning process.   

\subsection{Adversarial space}
CLIP architecture recognizes a wide variety of visual concepts in images and associate them with their names. Since, there is a large association between the image embedding and visual embedding space, a slight variation of the input image, can potentially cause the architecture to produce adversarial output. As discussed in \cite{tabacof2016exploring}, we want to explore the space of adversarial images in this architecture. Also, if possible and feasible, the idea of "restricting the directions of perturbations toward the existing words in the input embedding space" as mentioned in the paper \cite{sato2018interpretable} could also be explored for the CLIP architecture.

\subsection{Video Understanding/Action recognition/Video retrieval}
Action recognition is one of the actively researched areas in computer vision. This has led to collection of large sized datasets like kinetics \cite{smaira2020short}, something-something \cite{goyal2017something}, AVA etc. Most of the work is focused on classifying the actions given temporal and spatial localized clips as an  input \cite{feichtenhofer2019slowfast}. A different stream of research that works on non localized input is also being explored. This works by first generating proposals from the input videos and later classifying the proposals. Our work focuses on a different way of exploring video understanding, by looking at video as a continuous path in an image embedding space.
\section{Approach}

NLP models are usually trained on all the available text on the internet, we want to take advantage of this for open set video retrieval.
We do this by projecting videos into an visual embedding, which lie in the same joint embedding space as of text, Following an approach similar to what CLIP did for images.
We first extend the this clip image based model to videos. Given a video our approach projects it into the joint embedding space. We evaluate the quality of this model using action recognition on kinetics.
Then we remap the video classification model into retrieval setup. Given a textual query, our setup retrieves all the videos in the video database that match the textual query.

\subsection{Baseline}

\begin{figure}[h]
    \centering
    \includegraphics[width=1.0\textwidth]{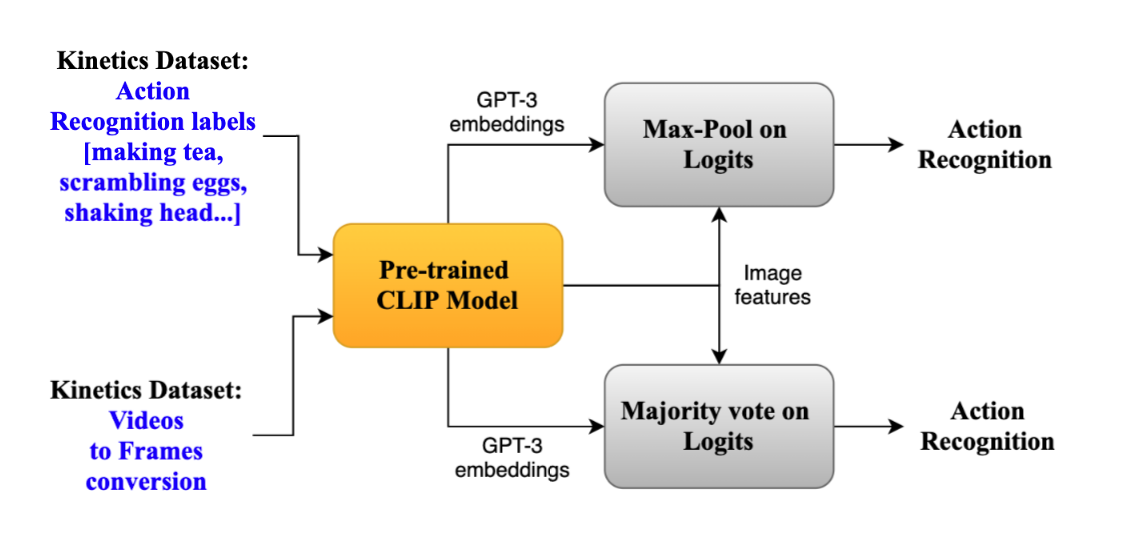}
    \caption{Proposed approah using GPT-3 for action recognition.}
    \label{fig:mask}
\end{figure}

Towards the first contribution of converting image based CLIP to video classification. We establish the following baselines
In first visual embeddings are extracted individually for the frames, followed by max pooling.
In second each image is individually classified, followed by majority vote to make the final prediction.

\subsection{Temporal fusion using LSTM}
In further experimentation, we want to project an entire video as an embedding in the joint embedding space.
Towards this we used a LSTM module that takes in series of image embeddings from the video and projects into a single embedding in the joint embedding space of GPT-3.
Dot product of textual embedding with visual embedding to get classification logits. Trained this on kinetics, and will show results in the upcoming slides.

\begin{figure}[h]
    \centering
    \includegraphics[width=1.0\textwidth]{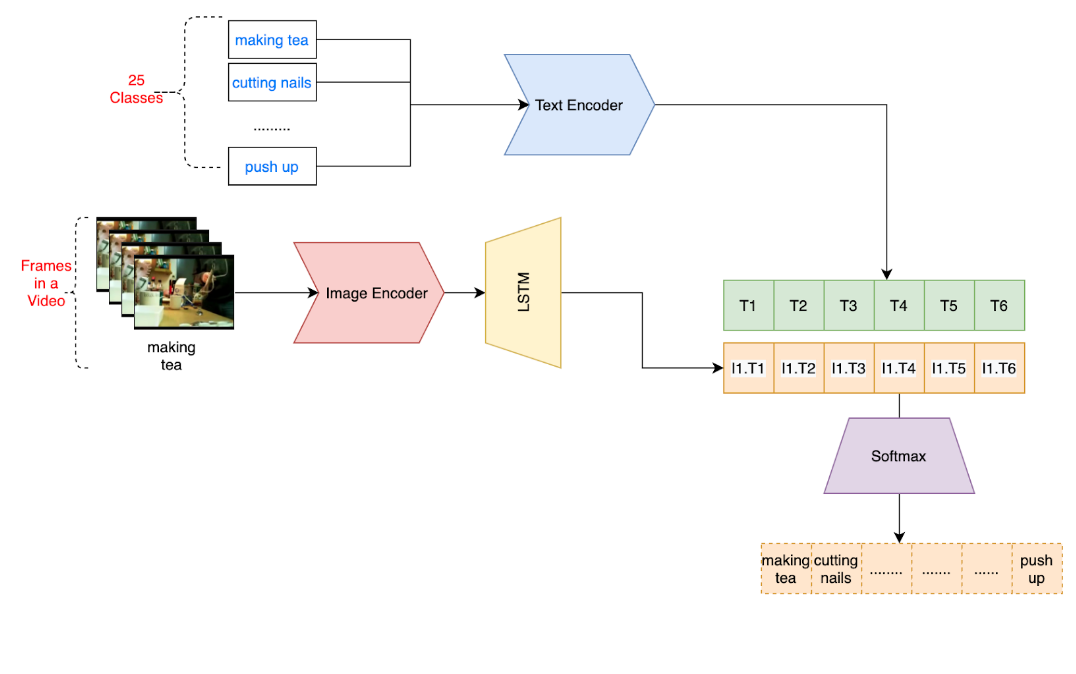}
    \caption{Proposed approach of CLIP with LSTM layer.}
    \label{fig:mask}
\end{figure}

\subsection{Temporal fusion using Transfomer}
We improved upon the previous approach by using a multi headed attention blocks to extract a visual embedding representing the input video.
\begin{figure}[!h]
    \centering
    \includegraphics[width=1.0\textwidth]{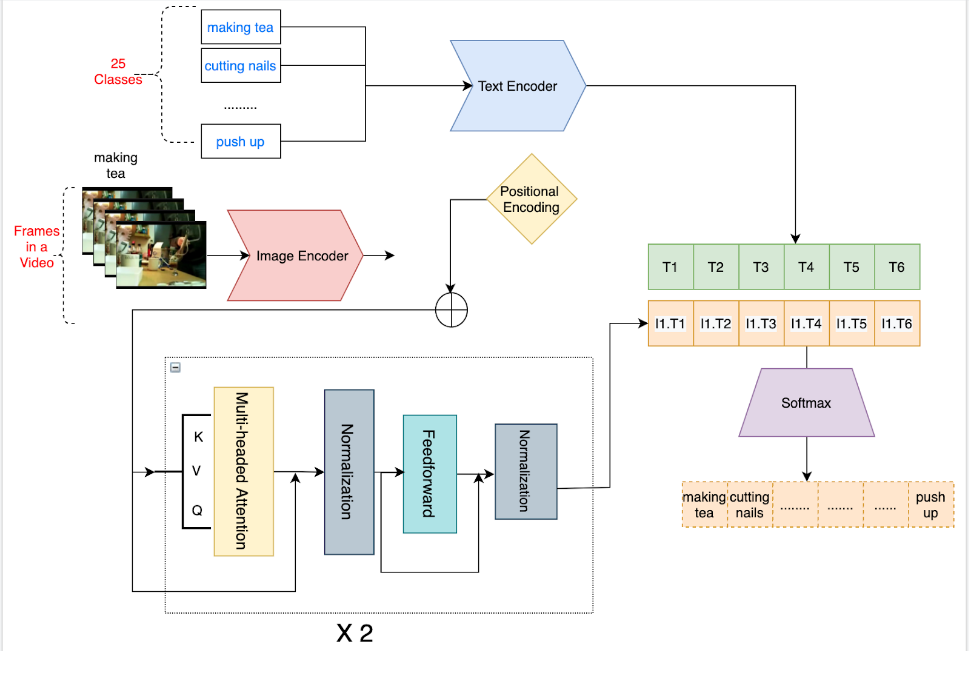}
    \caption{Proposed approach of CLIP with Multi-headed attention/Transformer Encoder.}
    \label{fig:mask}
\end{figure}

\subsection{Retrieval}

Finally we remapped the video classification model for retrieval. Using this pipeline given any textual query we retrieve all the videos that match the textual query.
To do this we use GPT-3 to first convert text query into an embedding.
Then compute the visual embedding for all the videos in the database using our classification model.(This step only needs to be done once and can reused across queries).
And use the dot product between the video embeddings and text embedding to get similarity score.
And returns the samples with highest similarity, dot product score.
\begin{figure}[h]
    \centering
    \includegraphics[width=1.0\textwidth]{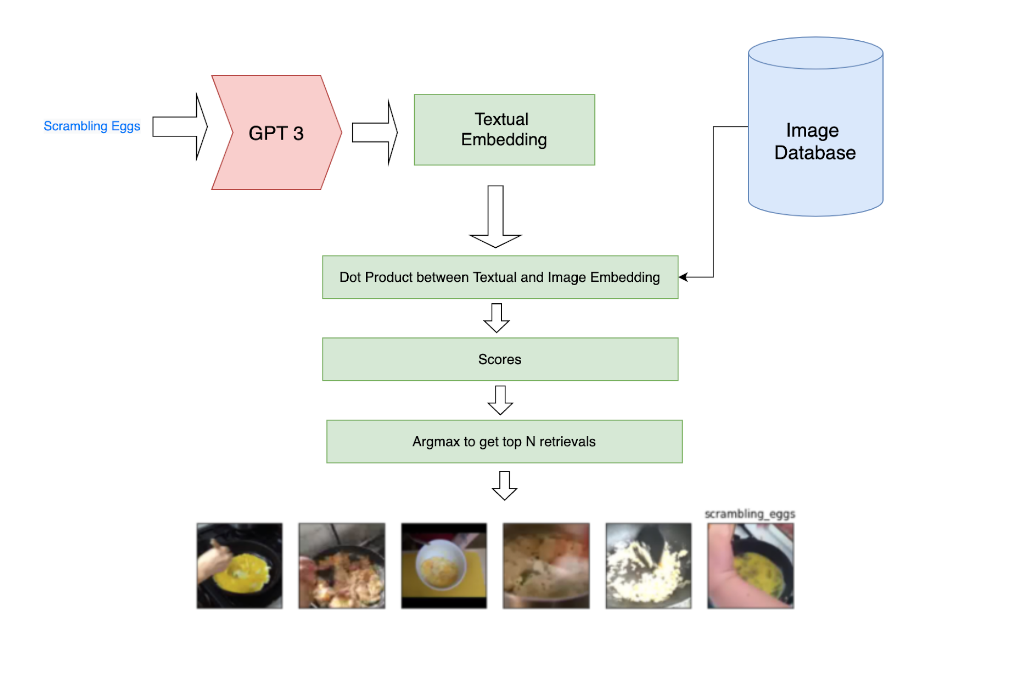}
    \caption{Video Retrieval.}
    \label{fig:mask}
\end{figure}

\section{Results}
The first task towards Video Retrieval requires the best action recognition model. Thus, we tried four different extensions of Image based CLIP Model to work for action recognition on videos. For the baseline we did a maxpool over all the frames in a video, and used CLIP Architecture to give the top voted class. Another, baseline architecture is taking majority votes over all the frames in a video. This gave a performance of 55.02

However, the above models did not take into account the temporal relationships between the frames in a video. We achieved that by adding LSTM/Transformer\cite{vaswani2017attention} to the existing CLIP Architecture, which gave a performance of 83.9

The original CLIP paper \cite{brown2020language}, takes a mid frame approach (taking middle frame in video and performing image-based classification on that) for action recognition and achieves an accuracy of only 53.4 percent.
Thus, adding temporal information improved the performance on video action recognition by around 32 percent.

The above model has been tested on only 25 classes of kinetics400 dataset, due to limited resources available on AWS in terms of video storage, frame storage and training. Also, we downsampled the video 100 frames per video, which might also have reduced the performance. In future, we would keep video length as a hyperparameter too.

We can see the loss curve of training and validation loss. The loss curve indicates overfitting, and the possible reason could be less number of classes and examples in the train dataset. We see similar behavior in both CLIP + LSTM and CLIP + Transformer Architecture.
\newpage
\subsection{Action Recognition}
\begin{figure}[!h]
    \centering
    \includegraphics[width=1.0\textwidth]{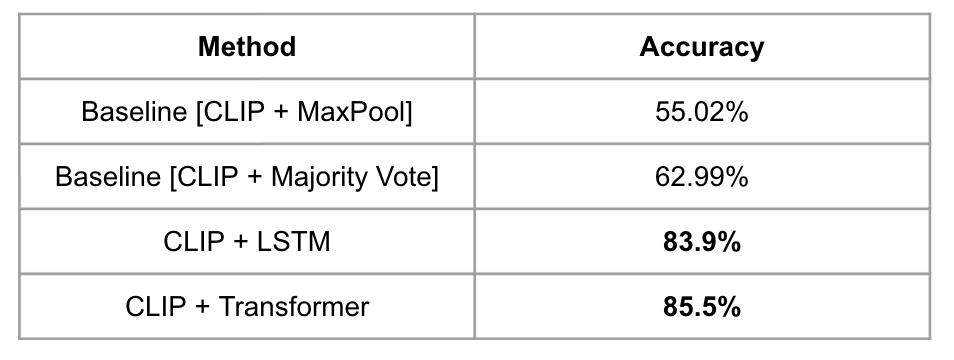}
    \caption{Accuracy comparisons.}
    \label{fig:mask}
\end{figure}

\begin{figure}[h]
    \centering
    \includegraphics[width=1.0\textwidth]{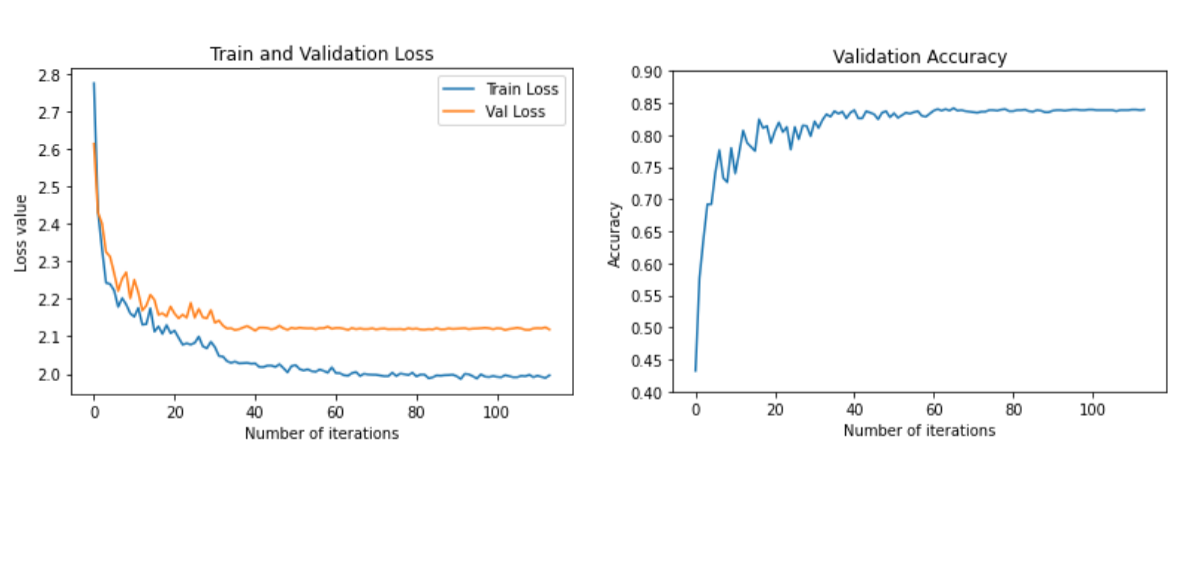}
    \caption{CLIP + LSTM ran over 25 classes of Kinetics 400 dataset for 50 epochs, using Adam Optimizer
and Ir = 1e-3 with 100 selected frames in a video.}
    \label{fig:mask}
\end{figure}
\begin{figure}[h]
    \centering
    \includegraphics[width=1.0\textwidth]{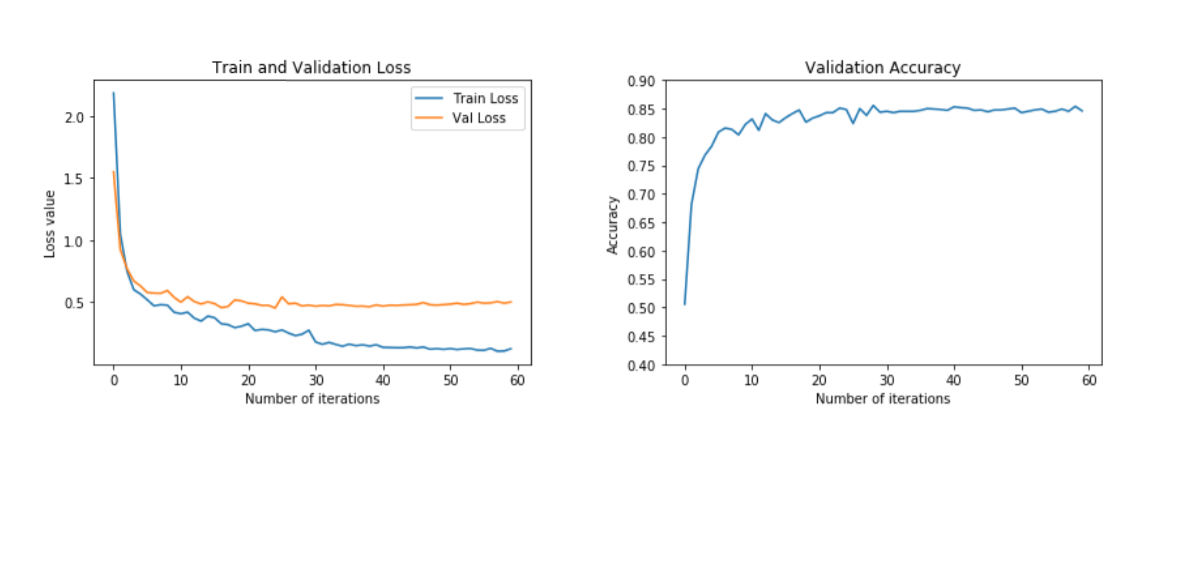}
    \caption{CLIP + Transformer ran over 25 classes of Kinetics 400 dataset for 50 epochs, using Adam
Optimizer and Ir = 1e-3 with 100 selected frames in a video.}
    \label{fig:mask}
\end{figure}
\newpage
\subsection{Retrieval}
\begin{figure}[!htbp]
    \centering
    \includegraphics[width=1.0\textwidth]{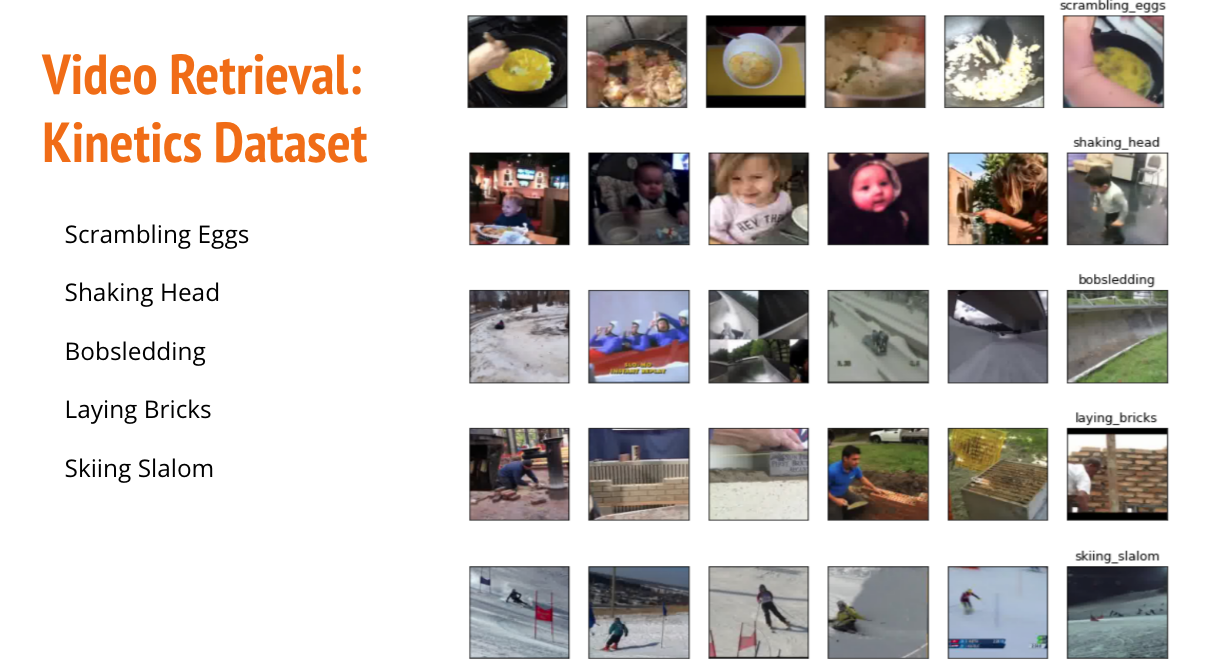}
    \caption{Top 6 video retrievals from Kinetics Dataset for the categories - Scrambling Eggs, Shaking Head, Bobsledding, Laying Bricks, Skiing Slalom.}
    \label{fig:mask}
\end{figure}

\begin{figure}[h]
    \centering
    \includegraphics[width=1.0\textwidth]{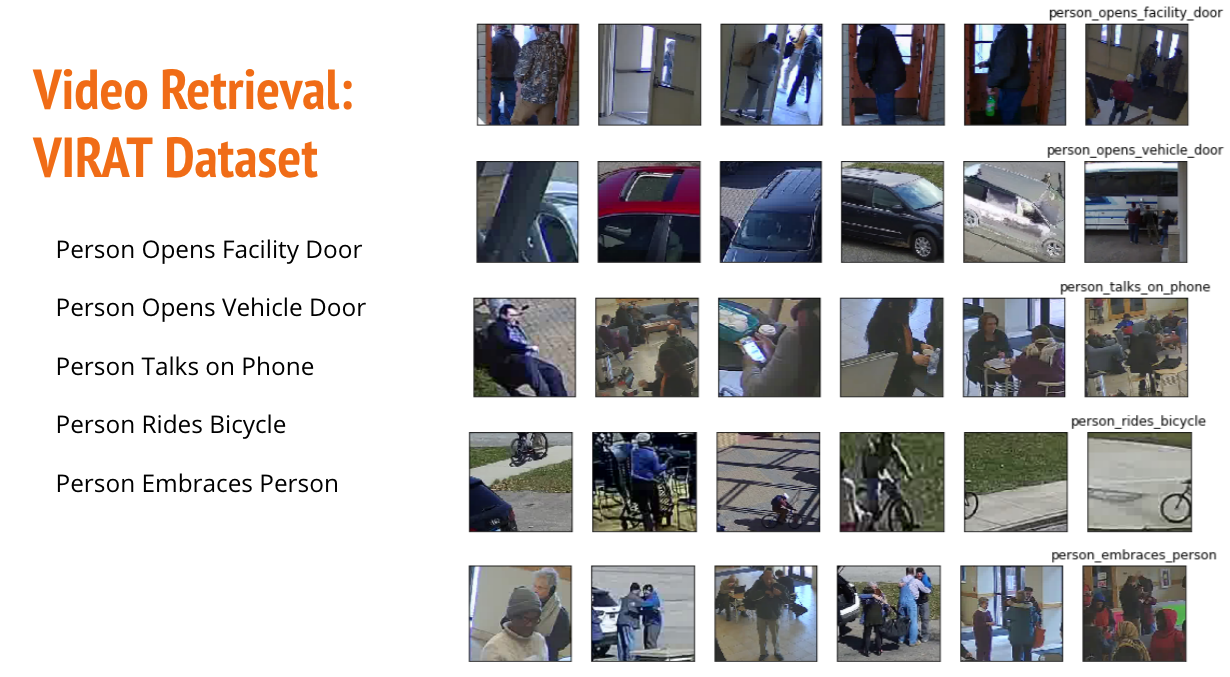}
    \caption{Top 6 video retrievals from the VIRAT dataset for various categories involving complex person object interaction.}
    \label{fig:mask}
\end{figure}

\newpage
\subsection{Emeddings}
\begin{figure}[h]
    \centering
    \includegraphics[width=1.0\textwidth]{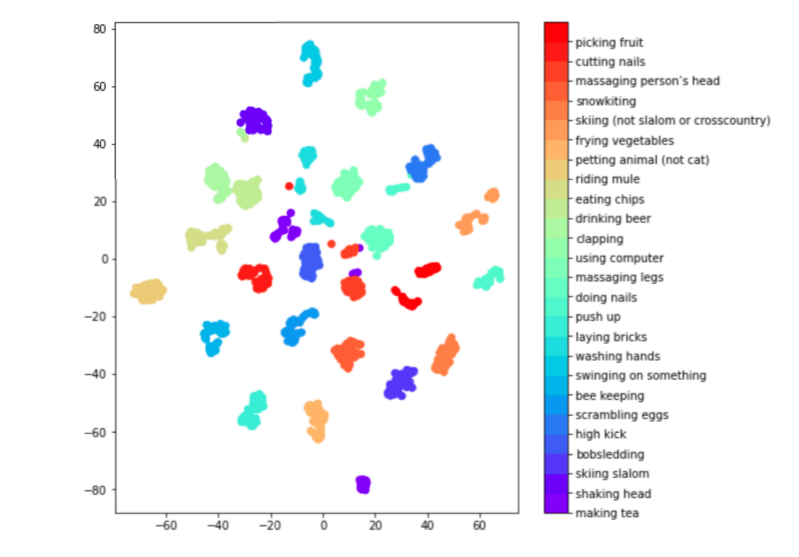}
    \caption{Embedding space visualisation for the various kinetics categories.}
    \label{fig:mask}
\end{figure}

\newpage
\section{Conclusion}
Conclusion
We extended the above trained models to Video Retrieval Task and got a remarkable result on image retrieval as indicated in the pictures. The way it works is that we calculate the text embedding of the entered query and get the embeddings of all the videos calculated using the above trained models (variations of CLIP) and get all the videos similar to the text embedding, by calculating the cosine similarity between text and video in the embedding space.

All the frames for a video cluster together as evident in the embedding space visualization (t-SNE - reduced from 1024 dimension to 2 dimension) of the video embedding corresponding to selected 25 classes.

The generalisation and performance of the model was done on two different datasets - Kinetics400 and VIRAT dataset as shown in Figures. Content Based Video Retrieval [egs: YouTube] is desirable because searches that rely purely on metadata are dependent on annotation quality and completeness. The presented architecture provides an efficient way of Text based Video Retrieval.

Also, our work is an extension of CLIP Model Action Recognition [RN50x16], where they have used Mid-frame level approach to get an accuracy of 53.4\%, while our method gives an improvement of 83.9\% [LSTM] and 85.5\% [Transformer] comparing to the SOTA on Kinetics 400 dataset 84.8\%.

\section{Future Work}
Our current work is based on classifying an action over the entire video. Using the same LSTM /Transformer model, we also wish to caption/add subtitle to each frame in the video with changing action.
For proof of concept, we have worked on 25 classes of Kinetics 400/700 dataset. In the future, we would be extending our work on the entire dataset with 400/700 classes.
We wish to see the performance on other complex datasets like MSR Action 3D and datasets like RareAct which contain unusual actions like “hammering a phone” and “drilling an egg’’. This will give us true performance of all the model over unusual/unseen actions too, which will validate the generalization of the model.
We also wish to explore the adversarial space of CLIP Models as different attacks like text patching and adversarial perturbation.

\bibliographystyle{plain}
\bibliography{neurips}
\end{document}